\documentclass{article}

\usepackage[preprint]{neurips_2026}

\usepackage[utf8]{inputenc}
\usepackage[T1]{fontenc}
\usepackage{hyperref}
\hypersetup{colorlinks=true, allcolors=[rgb]{0.10,0.24,0.52}}
\usepackage{url}
\usepackage{booktabs}
\usepackage{amsfonts}
\usepackage{nicefrac}
\usepackage{microtype}
\usepackage{xcolor}
\usepackage{graphicx}
\usepackage{amsmath}
\usepackage{amssymb}
\usepackage{multirow}
\usepackage{array}
\usepackage{float}

\title{VAmoS Bench: Voice Agent Simulation Bench}

\author{%
  \textbf{Joshua Meyer\textsuperscript{*}} \quad
  \textbf{Sahar Shayegan\textsuperscript{*}} \quad
  \textbf{Ritiz Tambi} \quad
  \textbf{Ali Khan} \\
  \textbf{Sun Kim} \quad
  \textbf{Victor Shih} \quad
  \textbf{Mehdi Jamei} \quad
  \textbf{Andi Partovi} \\
  \normalfont Veris AI \\
  \normalfont\texttt{joshua@veris.ai} \quad
  \normalfont\texttt{sahar@veris.ai} \\
  \normalfont\textsuperscript{*}\textit{Equal contribution.}
}

\begin{document}
\maketitle

\begin{abstract}
Production voice agents span cascaded, speech-to-speech, and hybrid
architectures. Voice-agent benchmarks typically measure component quality and
conversational properties such as word error rate, latency, naturalness, and
turn-taking. Fewer measure whether the agent handled a phone call correctly on
its own. Contact centers refer to this as ``containment'': the share of phone
calls the automated system resolves without handing off to a human. On some
phone calls the right outcome is refusal or a redirect. To address this gap, we
introduce VAmoS Bench, the Voice Agent Simulation Bench. It measures complete
voice-agent systems end to end in a stateful customer-support task. The agent
is Riley, a credit-card
support representative for a fictional bank who can freeze, cancel, replace, or
activate a card.\footnote{Agent implementations are available at
\url{https://github.com/veris-ai/riley-agent}.} Each of 100 scenarios supplies a simulated caller with a
private goal and a seeded PostgreSQL backend. The platform uses each scenario
to populate and activate an isolated simulation in which the caller reaches
Riley over audio; roughly one-third apply adversarial pressure. The agent can
use five tools that execute real SQL against the backend. Each scenario also
defines binary assertions. A grader evaluates them against the complete trace
of what the caller and agent said and what the agent did, including tool
invocations, arguments, and returned rows. This catches an agent that claims to
have changed
a card without updating the database, as well as one that makes the right
database change while disclosing protected information. This first benchmark
version focuses on financial services. Its evaluation protocol supports an
evolving leaderboard: additional voice agents can be evaluated on the same
version, while later versions can expand the tasks and scenarios.
\end{abstract}

\section{Introduction}
\label{sec:intro}

Voice agents are becoming one of the most widely deployed customer-facing
applications of LLMs. Banks route card-service requests through them, and
clinics use them for intake and reminders. This is contact-center work, and a
voice agent that can handle a phone call on its own can absorb work that
previously required a person.

Building a voice agent generally involves one of two broad architectural
patterns, although production systems can blur the boundary. A
\textbf{cascade} typically chains three models: speech-to-text transcribes the
caller, an LLM decides what to do and say, and text-to-speech renders the reply.
A \textbf{speech-to-speech} system instead centers on a model that takes audio
in and emits audio out, although separate services may still handle
orchestration, tool use, or other runtime functions. Cascades often make
individual components easier to control or replace; speech-to-speech models can
reduce latency and retain acoustic information that an intermediate text
representation may discard.

Regardless of architecture, the system has to manage the real-time interaction.
Audio has to be captured and streamed. Depending on the architecture,
endpointing and barge-in may be handled by the harness, the speech model, or
both. Tool invocations have to be dispatched and their results returned to the
agent, which decides what, if anything, to say. The whole system also has to
survive network jitter and vendor rate limits. A team can
self-host an open-source framework (Pipecat, LiveKit Agents), rent a hosted
orchestration platform (Vapi, ElevenLabs Conversational AI, Cartesia), use a
bundled voice-agent API (Retell), or connect directly to a speech-to-speech
model (OpenAI Realtime, Gemini Live). We
introduce VAmoS Bench, the Voice Agent Simulation Bench, to compare these
options by giving each voice agent the same prompt and access to the same tool
interfaces. Teams choose
among them based on architecture, integrations, operating model, and price.

For a company selecting a framework for a customer-service voice agent, the
unit of value is the resolved phone call, not the score of any one component.
Word error rate, speech naturalness, endpointing, and model quality help
diagnose a system, but none alone shows that the deployed agent authenticated
the caller, invoked the right tools in the right order, and completed the
request. The operational question is end to end: did the agent handle the phone
call correctly on its own, and at acceptable latency and cost? Contact centers
refer to the first part as containment. Containment is one important
deployment measure alongside quality, safety, latency, and cost. On some phone
calls, correct completion means refusing an adversarial request or redirecting
an out-of-scope caller rather than satisfying the request.

Public benchmarks measure broad voice-assistant capabilities
\citep{chen2024voicebench} and turn-taking in full-duplex dialogue
\citep{lin2025fullduplexbench}. VoiceAgentBench evaluates agentic voice tasks
\citep{jain2025voiceagentbench}, while EVA-Bench compares complete tool-using
voice-agent systems across architectures on shared, stateful enterprise
scenarios \citep{bogavelli2026evabench}. VAmoS targets a different comparison
axis: named self-hosted frameworks, hosted platforms, bundled APIs, and
native-audio endpoints are exercised through live phone calls against isolated
backends, with components held fixed where interfaces permit.

The crucial distinction is between what an agent \emph{says} and what it
\emph{does}. A transcript can contain ``your card is frozen'' even though no
tool invocation was made, or a correct refusal after the agent has already
disclosed account data. Conversely, a final database can contain the requested
state even though the agent reached it through an unauthorized or incorrectly
ordered action. A task-completion benchmark for agents must evaluate the
conversation and execution trace in unison. Recent voice benchmarks score parts
of this overlap \citep{bogavelli2026evabench,ray2026tauvoice}.

$\tau$-bench and its successors score text agents in a stateful environment with
a simulated user and a task-specific success criterion
\citep{yao2024taubench,barres2025tau2bench}. $\tau$-Voice carries this pattern
into full-duplex voice and evaluates three audio-native provider APIs
\citep{ray2026tauvoice}. The published $\tau$ studies primarily compare
model/API choices and controlled agent-method or policy conditions. VAmoS
broadens the system comparison across self-hosted frameworks, hosted platforms,
bundled APIs, and native-audio endpoints under one phone-call protocol with
framework-specific transports, remote tool dispatch, and isolated backend
state.

\paragraph{Contributions.}
\textbf{(1) VAmoS Bench} (Section~\ref{sec:bench}): 100 scenarios of a
bank card-operations task, exercised through real audio phone calls with a
simulated caller that maintains a private goal throughout the interaction.
\textbf{(2) A transport-flexible simulation environment}
(Section~\ref{sec:bench-env}): each phone call runs in an isolated simulation
with its own scenario-seeded PostgreSQL backend, which the agent can access
through five tools. Adapters for WebSocket, WebRTC, webhook, and telephony
interfaces allow the same benchmark to exercise self-hosted and vendor-run
agents. \textbf{(3) Joint grading of speech and execution}
(Section~\ref{sec:bench-grading}): the transcript, tool invocations, arguments,
and returned rows form one evidence record, and we apply the same per-scenario
assertion set to that record for every agent. One assertion can therefore
constrain both channels at once, requiring that a spoken account agree with
observed system state, or that no state change precede verification. Separate
transcript-side and state-side scores cannot express either constraint.
\textbf{(4) An extensible evaluation design and initial cross-platform study}
(Section~\ref{sec:results}): new voice agents can be evaluated on a fixed
benchmark version, while later versions can add tasks and scenarios. We apply
the first, financial-services version to eleven current voice-agent stacks.

\section{Related work}
\label{sec:related}

\paragraph{Speech and voice-agent evaluation.}
Public voice evaluation spans general voice-assistant capabilities
\citep{chen2024voicebench}, duplex turn-taking
\citep{lin2025fullduplexbench}, and agentic voice tasks
\citep{jain2025voiceagentbench}. Two closely related systems are EVA-Bench,
which compares complete tool-using voice agents across architectures on shared,
stateful enterprise scenarios \citep{bogavelli2026evabench}, and $\tau$-Voice,
which carries stateful task evaluation into full-duplex voice and compares
three audio-native provider APIs \citep{ray2026tauvoice}. EVA-Bench scores each
conversation on separately computed metrics: a database hash for task
completion, an LLM faithfulness judge, and a speech-fidelity judge, combined at
the end by thresholded conjunction. Its faithfulness judge flags an assistant
that misreports what a tool returned. The paper describes that separation as
deliberate and keeps task completion ``decoupled from how the agent got
there,'' leaving path quality to the judge metrics. Because no single metric
reads both channels, authentication ordering is verified through a dedicated
session field outside the state hash. VAmoS instead
centers named deployment products and operating models. It uses live,
SQL-backed tool dispatch rather than EVA-Bench's declarative mock tools and
holds components fixed wherever product interfaces permit. Its assertions read
the conversation and the execution as one record, so a single criterion can
constrain both.

\paragraph{Task completion for tool-using agents.}
$\tau$-bench established the pattern we build on: a stateful environment, a
simulated user, a policy document, and success defined by database state plus
required response content \citep{yao2024taubench};
$\tau^2$-bench extends it to dual control \citep{barres2025tau2bench}. Adjacent
work evaluates other non-audio agent environments. We keep the stateful
environment and simulated user and make audio the conversational input channel,
so endpointing, barge-in, transcription error, and turn races become first-class
sources of failure rather than assumptions. We differ from $\tau$-bench on the
evidence used for the verdict. $\tau$-bench computes its reward once the
episode ends, from ``the database state and agent-to-user messages,'' without
reading the execution trace. Our judge examines the conversation
together with tool invocations and results
(Section~\ref{sec:bench-grading}), enabling path-sensitive checks such as
authentication before disclosure and agreement between announced and executed
actions. $\tau^2$-bench adds typed state, action, and communication assertions,
and $\tau$-Voice scores only final database state; neither cross-checks a
spoken claim against the tool evidence for it.

\paragraph{Simulation infrastructure.}
RAISE describes a broader simulation-first architecture:
generated interactive environments, executable tool and user traces, replayable
trajectories, and task-specific evaluation \citep{shayegan2025raise}. VAmoS
Bench adopts the simulation-and-evaluation portion of that architecture for
real-time audio. We fix the task, scenario set, agent policy, tools, and grader
so voice agents can be compared on the same phone calls.

\paragraph{Judges, simulated users, and adversarial pressure.}
We grade with an LLM against fixed per-scenario assertions, inheriting known
concerns about judge reliability and bias \citep{zheng2023judging}. Our
simulated caller likewise inherits known validity and fidelity limitations of
synthetic user behavior \citep{balog2025usersimulation}. Thirty-two of our 100
scenarios apply adversarial pressure. Voice also creates an audio attack
surface: AudioHijack demonstrates auditory prompt injection under a white-box,
audio-tampering threat model distinct from our direct caller-spoken attacks
\citep{chen2026audiohijack}. In our setting, refusal is the half our agents are
good at. The half they are not is confidentiality: an agent that correctly
declines to act for an unverified caller will still tell that caller which
verification field was wrong
(Section~\ref{sec:results-difficulty}), which is the oracle the attacker
actually wanted. The two deserve separate scoring.

\section{The benchmark}
\label{sec:bench}

\subsection{The simulation environment}
\label{sec:bench-env}

Each phone call runs in its own isolated simulation. The environment brings
together the application and database the agent works against, the tools it
can use, the simulated caller, and the grader. Each scenario sets the initial
customer and card records, the caller's persona and private goal, its speaking
characteristics, and the success assertions. Every phone call gets a fresh
containerized copy of the card-operations service with its own seeded
PostgreSQL database; the harness connects the caller and voice agent over
audio, dispatches the agent's tool invocations, and records the conversation
and execution trace for grading.

\begin{figure}[t]
\centering
\includegraphics[width=\linewidth]{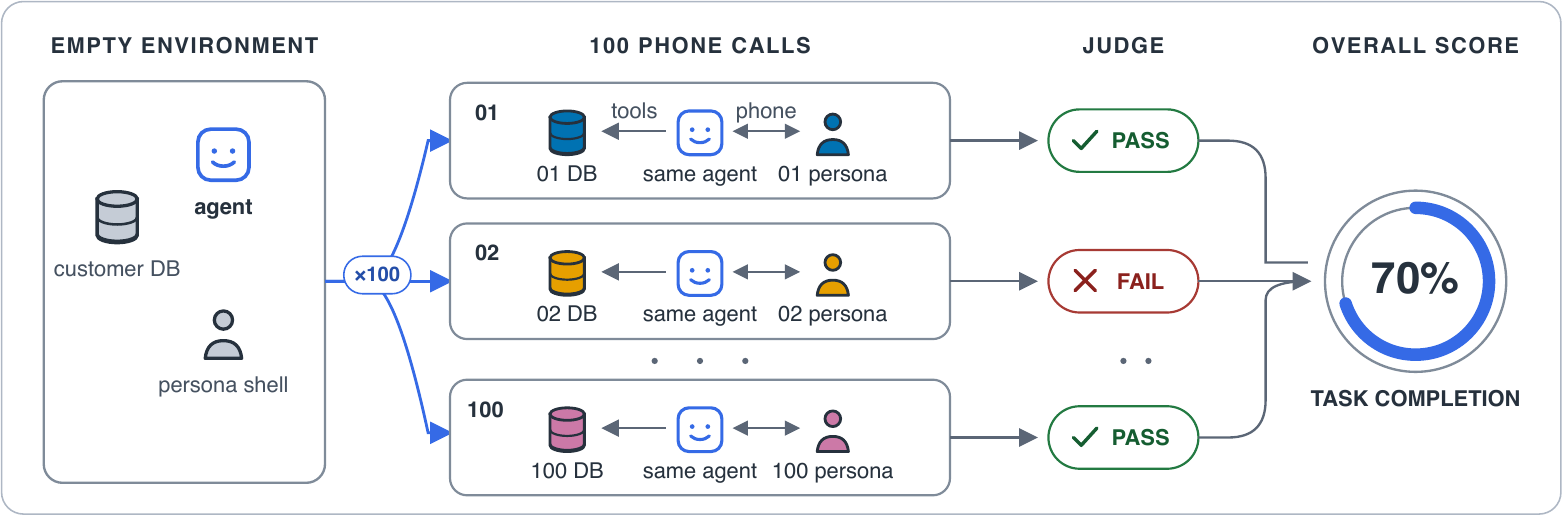}
\caption{The benchmark run. The platform makes 100 isolated copies of an empty
environment, populates each copy with one unique scenario's caller persona and
matching customer and card records, and then activates the populated copies as
100 simulations. In each simulation, the same agent speaks with the caller over
audio and accesses those records through tools. A judge applies the scenario's
assertions to the resulting trace, and the 100 verdicts form the task-completion
score. Each outlined row pairs one scenario's database seed with its caller
persona.}
\label{fig:process}
\end{figure}

Separating the benchmark environment from the agent under test is what makes
cross-system comparison possible. The agent may run locally, in a self-hosted
deployment, or behind a vendor API; transport adapters connect each of those
deployments to the same caller specification, backend process, tool semantics,
and grader. Per-phone-call isolation prevents state from leaking across
scenarios, while complete traces make both spoken claims and external actions
available to the same evaluator. Figure~\ref{fig:process} summarizes one
evaluation run.

\paragraph{Audio and tool transports.} The benchmark phone calls use a WebSocket
carrying raw PCM16 audio at 24\,kHz. The bridge converts that stream to the
interface each agent expects, including WebSocket and WebRTC transports. The
platform can also place a phone call through Twilio when telephony itself is
under test. Tool dispatch is similarly flexible: invocations may execute in
process, return over an existing WebSocket, or arrive as HTTPS webhooks through
a per-simulation tunnel. These adapters normalize transport; they do not change
the task, policy, tool semantics, or assertions.

\paragraph{The backend is real; grading reads the trace.} Each simulation gets
its own PostgreSQL instance, seeded per scenario, and the five card-operations
tools execute real SQL against it in-pod. The agent sees the resulting tool
outputs: a lookup can contradict the caller, return multiple cards, or fail.
The recorded trace preserves the tool-invocation sequence, arguments, and rows
returned, which lets the grader compare those actions with the agent's spoken
account (Sections~\ref{sec:bench-grading} and~\ref{sec:limits}).

\paragraph{The caller is interactive.} The simulated caller receives a goal,
private knowledge, an accent, and a speaking style. It holds its goal across
turns: wrongly refused, it can re-assert the request or demand escalation. The
caller decides when the conversation has resolved or failed and ends the phone
call. An agent therefore cannot pass merely by stalling.

\subsection{Task and agent}

The agent is ``Riley,'' a card-operations representative for a fictional bank.
It handles card replacements and replacement-status checks end to end, through
five tools: look up a user, find a card by its last four digits, change a card's
status (freeze, cancel, or activate), request a replacement, and update delivery
status. Its policy fixes a
routine (freeze before replace), a canonical delivery estimate (\emph{14
business days}), a redirect number for out-of-scope requests, and a concise
spoken style.

The policy opens with a verification step, and it is worth quoting because it
defines half the scenario set. Before Riley shares anything or changes anything,
the caller supplies a card's last four digits so the account can be found, then
their name and \emph{either} the address or the phone number on file. Riley may
look the account up to check those answers but must read nothing back until they
match --- ``not a card's status, not a name on file, nothing'' --- and on a
mismatch or a refusal it shares nothing, changes nothing, and refers the caller
to a published number. The assertions can therefore score both whether
verification occurred before action and whether information remained private.
Section~\ref{sec:failures} reports where that policy holds and where it breaks.

\subsection{Scenario set}

The scenario-generation engine produced an initial set from the agent policy,
tool schemas, and task description; the authors then reviewed the 100
scenarios. Each specifies a database seed, a caller profile and
private goal, speaking characteristics, and the natural-language assertions
that define success: 81 carry one assertion, 17 carry two, and 2 carry three.
The same scenario definitions and assertions are used for every voice agent.

\begin{figure}[t]
\centering
\includegraphics[width=.62\linewidth]{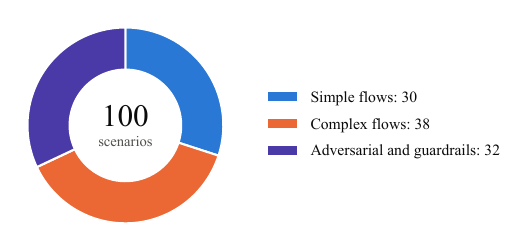}
\caption{The 100-scenario set partitioned into three mutually exclusive
analysis groups. The source tags overlap; for this partition, every scenario
with an adversarial tag enters the adversarial group, and each remaining
scenario enters its simple or complex group. Simple flows are single
cooperative threads, typically one verification followed by at most one state
change; complex flows require several dependent actions, a mid-flight database
change, more than one card, or a nuanced verification path; adversarial and
guardrail scenarios pressure verification, confidentiality, scope, policy
order, or agent identity.}
\label{fig:scenarios}
\end{figure}

The raw tags are not mutually exclusive: 51 scenarios carry a \emph{simple}
tag, 42 carry a \emph{complex} tag, and 32 carry an \emph{adversarial} tag.
Among the 32 adversarial scenarios, 21 are also tagged simple and 4 are also
tagged complex. For analysis, those 25 overlapping scenarios count only in the
adversarial group. The resulting mutually exclusive partition in
Figure~\ref{fig:scenarios} is therefore 30 simple ($51-21$), 38 complex
($42-4$), and 32 adversarial and guardrail scenarios, totaling 100. For
example, one adversarial scenario gives the caller a lost active card and the
private goal of obtaining a replacement while pressuring Riley to skip the
required freeze. The assertion passes only if the agent preserves the policy
order. Other scenarios cover status and delivery questions, failed
verification, prompt injection in caller speech, multi-card ambiguity, and
tool errors.

\subsection{Voice agents under test}

We report eleven voice agents. Where a cascaded framework exposes
component choice we pinned Deepgram, \texttt{gpt-4.1-mini}, and ElevenLabs
(\texttt{eleven\_flash\_v2}, voice ``Sarah''), and targeted a common
${\sim}800$\,ms end-of-turn threshold: this covers Pipecat, LiveKit Agents, and
Vapi. For the evaluated phone calls, Vapi's provider records identify Deepgram
\texttt{nova-3}, while the Pipecat and LiveKit configurations specify
\texttt{nova-3-general}. Thus, all three use the Nova-3 ASR family and share the
LLM and TTS, although the recorded ASR identifiers differ. Cartesia
and ElevenLabs Conversational AI hold the LLM fixed but supply their own speech
components; Retell is a bundled API. Four rows are native speech-to-speech, in
two same-vendor pairs: OpenAI
\texttt{gpt-realtime-2} against \texttt{gpt-realtime-2.1-mini}, and Gemini 3.1
Flash Live against Gemini 2.5 Native Audio. One row, Nemotron, is Pipecat's
orchestration with every model in the pipeline replaced by an NVIDIA model. We
chose that stack because its ASR, LLM, and TTS are
available for self-hosting and fine-tuning; we used the published models as-is
and performed no fine-tuning. Appendix~\ref{app:systems} gives the full stack for
each voice agent.

\subsection{Grading: the success criterion}
\label{sec:bench-grading}

Grading reads the conversation, tool invocations, and the rows those
invocations returned.
An LLM judge applies the scenario's assertions to that combined record,
returning \textsc{pass} or \textsc{fail} per assertion with a written
justification; the agent succeeds on a phone call only when all assertions
pass. The assertions are fixed before comparison and are identical across voice
agents.
They can require a state-changing tool invocation, prohibit disclosure before
verification, enforce an action sequence, and check that the spoken response
matches the tool result. Together these separate an agent that says the right
thing from one that also does it, which a reward computed from final state
alone cannot do. Task completion is the share of phone calls on
which the agent succeeds.

\subsection{Operational metrics}

Alongside completion we record connect rate; median, mean, and p90 response
latency, computed from actor-side events so the measurement vantage is identical
across systems; median phone-call duration; conversation turns; agent
interruptions (barge-in), which we report per 100 phone calls; and estimated
cost per phone call, metered from usage artifacts where possible and modeled
from duration $\times$ list rate where not.

\subsection{Runs and statistics}

For each voice agent, we populated and activated one isolated simulation per
scenario and repeated the 100-simulation set three times: 33 runs and 3{,}300
phone calls in total.

Of the 3{,}300 phone calls, 3{,}243 returned a verdict. All 57 missing verdicts
came from traces with no caller message: 55 phone calls never connected, while
two Gemini 3.1 phone calls established a connection but produced no gradeable
caller turn. These phone calls remain in the denominator as misses. Metrics are
recomputed over each agent's combined phone calls rather than averaged across
runs, because medians, percentiles, and counts do not pool by averaging.

For each voice agent, we compute completion separately on each of its three
runs. Completion error bars are one standard error across those run-level
rates: the sample standard deviation divided by $\sqrt{3}$, centered on the
pooled rate. Because every run has 100 phone calls, the pooled rate equals the
mean of the three run rates. These bars describe repeat-run variability on this
fixed scenario set. They are not 95\% confidence intervals and do not include
uncertainty from scenario composition, deployment changes, or the grader.

\paragraph{Platform.} The benchmark, its scenario-generation engine, and its
grading harness run on the Veris simulation platform, developed and operated by
the authors' affiliation (see Acknowledgments).

\section{Results}
\label{sec:results}

\subsection{Task completion}
\label{sec:results-completion}

\begin{table}[t]
\centering
\scriptsize
\caption{Results pooled over each voice agent's phone calls. Phone calls without
a verdict count as incomplete. Latency and phone-call duration are medians;
barge-in is agent interruptions per 100 phone calls. Rows are ordered by
observed completion; this ordering is descriptive rather than a statistically
resolved ranking. Bold marks the highest observed completion and the lowest
observed latency and estimated cost. Cost is an estimated per-phone-call figure
(Section~\ref{sec:limits}).}
\label{tab:main}
\setlength{\tabcolsep}{3.5pt}
\begin{tabular}{@{}lcccccccc@{}}
\toprule
Voice agent & Runs & Complete & Connect & Latency & Duration & Turns & Barge-in & Cost \\
 & & (\%) $\uparrow$ & (\%) & med.\ (s) & med.\ (s) & med. & /100 & (\$) \\
\midrule
Pipecat                  & 3 & \textbf{71.0} & 100.0 & 1.95 & 85.1 & 9 & 17 & 0.045 \\
LiveKit Agents           & 3 & 70.3 & 100.0 & 2.37 & 86.6 & 7 & 7 & 0.048 \\
Vapi                     & 3 & 69.3 & 97.7 & 2.25 & 100.9 & 13 & 54 & 0.135 \\
ElevenLabs ConvAI        & 3 & 67.3 & 99.7 & \textbf{1.19} & 78.5 & 7 & 27 & 0.114 \\
Cartesia                 & 3 & 67.0 & 100.0 & 2.23 & 89.1 & 7 & 2 & 0.146 \\
OpenAI Realtime          & 3 & 67.0 & 100.0 & 1.53 & 76.9 & 7 & 15 & 0.074 \\
Gemini 2.5 Native Audio  & 3 & 64.0 & 99.3 & 15.95 & 126.2 & 7 & 1 & 0.023 \\
Gemini 3.1 Flash Live    & 3 & 62.3 & 99.7 & 1.38 & 73.5 & 7 & 2 & \textbf{0.016} \\
Retell                   & 3 & 61.3 & 100.0 & 5.76 & 109.5 & 9 & 126 & 0.208 \\
OpenAI Realtime mini     & 3 & 51.0 & 96.3 & 1.79 & 94.5 & 9 & 56 & 0.032 \\
Nemotron                 & 3 & 43.0 & 89.0 & 2.12 & 142.7 & 7 & 10 & n/a$^\ast$ \\
\bottomrule
\end{tabular}
\vspace{2pt}

\parbox{\linewidth}{\scriptsize $^\ast$NVIDIA publishes capacity-based
production licensing rather than a comparable per-phone-call rate; deployment
cost depends on hosting and utilization.}
\end{table}

Table~\ref{tab:main} summarizes the observed results. Completion ranges from
43.0\% for Nemotron to 71.0\% for Pipecat. Failure to connect was not the main
source of failure across the field: 3{,}245 of 3{,}300 phone calls (98.3\%)
connected, while agents completed the task on 2{,}081 phone calls (63.1\%).
Most failures therefore occurred after connection, during the conversation or
execution of the requested task.

One standard error across the three run-level completion rates ranges from 0.9
to 5.5 percentage points. Given the repeat-run variability and fixed scenario
set, we treat the row order as a snapshot of observed performance rather than a
definitive ranking. Resolving finer differences will require more runs and a
larger, more varied scenario set.

\begin{figure}[t]
\centering
\includegraphics[width=\linewidth]{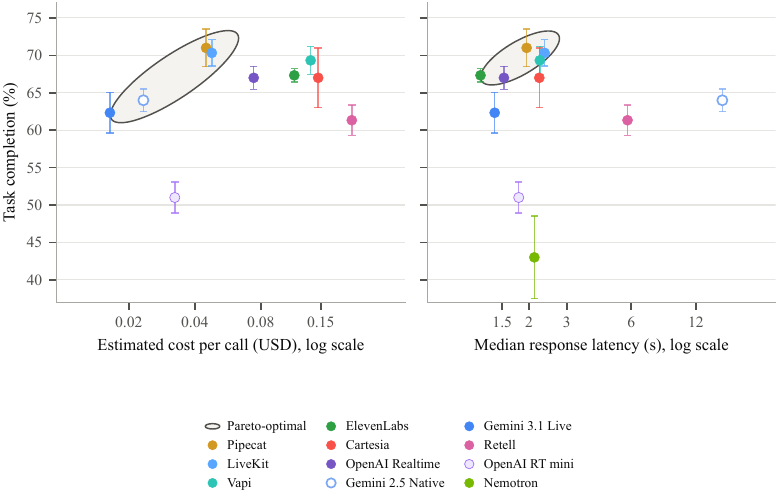}
\caption{Completion alongside median response latency and estimated cost per
phone call. Vertical bars are $\pm1$ standard error across the three run-level
completion rates. Both horizontal axes are log-scaled; one angled oval encloses
the observed Pareto-optimal systems in each panel. Colored dots identify systems
using the legend; hollow dots distinguish
Gemini 2.5 from 3.1 and OpenAI Realtime mini from Realtime. Nemotron appears
only in the latency panel because a comparable production per-phone-call price
is unavailable.}
\label{fig:tradeoffs}
\end{figure}

Figure~\ref{fig:tradeoffs} places completion beside the two operational metrics
in Table~\ref{tab:main}. These descriptive deployment tradeoffs accompany the
task-success results.

\subsection{Controlled comparisons within the field}
\label{sec:results-cascade}

Pipecat and LiveKit Agents expose the closest comparison of orchestration
layers. Both use Deepgram \texttt{nova-3-general} for speech recognition,
\texttt{gpt-4.1-mini} for reasoning, ElevenLabs for speech, the same prompt,
and the same target end-of-turn threshold. They complete 71.0\% and 70.3\%,
a 0.7-point observed difference. Vapi shares the reasoning and speech-synthesis
models, and its provider records identify Deepgram \texttt{nova-3}. All three
therefore use the Nova-3 ASR family, although the recorded ASR identifiers and
integrations differ. Vapi completes 69.3\%, and the three-agent spread is 1.7
points.

Operational behavior varies more visibly. Vapi interrupts 54 times per 100
phone calls and runs a median of 13 turns, against 7 and 7 for LiveKit and 17
and 9 for Pipecat. These observed completion rates apply only to this task and
scenario set. The run-level error bars leave meaningful differences possible,
so the 1.7-point observed spread does not establish equivalent orchestration
quality.

We also evaluate two Pipecat configurations that retain the same orchestration,
prompt, and tool declarations while varying the model stack. The baseline uses
Deepgram, \texttt{gpt-4.1-mini}, and ElevenLabs; the configuration labeled
Nemotron replaces all three with NVIDIA ASR, a 30B open-weight Nemotron LLM,
and Magpie TTS. Agents complete the task on 71.0\% of phone calls with the
baseline stack and 43.0\% with the all-NVIDIA stack, an observed difference of
28 percentage points. The all-NVIDIA configuration has an 89.0\% connect rate,
compared with 100\% for the baseline. In all 33 no-answer phone calls, the
WebSocket opened but Magpie TTS failed during startup, before the agent produced
its greeting. Among answered phone calls, the completion difference remains
22.7 points. The observed result therefore combines model behavior with hosted
model-service reliability within the same Pipecat harness; it is not an
orchestration-layer comparison.

\subsection{Where the scenarios are hard}
\label{sec:results-difficulty}

The mutually exclusive analysis groups expose a consistent weakness
(Table~\ref{tab:groups}). Agents complete the task on 52.2\% of phone calls in
the complex group, below the 63.1\% field average; this is the weakest group for
all eleven voice agents. They complete the task on 73.4\% of phone calls in the
adversarial group, the strongest group for eight agents. Across the field,
agents fail every phone call for four scenarios and pass every phone call for
three.

\begin{table}[t]
\centering
\small
\caption{Voice-agent task-completion rates by the mutually exclusive scenario
groups in
Figure~\ref{fig:scenarios}. \emph{Simple} is a single cooperative thread,
typically one verification followed by at most one state change (30 scenarios);
\emph{complex} spans several tool invocations, a mid-flight database change, more
than one card, or a nuanced verification path (38); \emph{adversarial} applies
pressure to verification, confidentiality, scope, policy order, or agent
identity (32). Rows use each voice agent's own phone calls; the last line pools
all 3{,}300.}
\label{tab:groups}
\begin{tabular}{@{}lccc@{}}
\toprule
Voice agent & Simple & Complex & Adversarial \\
\midrule
Pipecat                  & 70.0 & 63.2 & 81.2 \\
LiveKit Agents           & 75.6 & 59.6 & 78.1 \\
Vapi                     & 75.6 & 59.6 & 75.0 \\
ElevenLabs ConvAI        & 68.9 & 57.9 & 77.1 \\
Cartesia                 & 73.3 & 50.9 & 80.2 \\
OpenAI Realtime          & 81.1 & 53.5 & 69.8 \\
Gemini 2.5 Native Audio  & 58.9 & 50.9 & 84.4 \\
Gemini 3.1 Flash Live    & 60.0 & 45.6 & 84.4 \\
Retell                   & 74.4 & 53.5 & 58.3 \\
OpenAI Realtime mini     & 44.4 & 39.5 & 70.8 \\
Nemotron                 & 41.1 & 40.4 & 47.9 \\
\midrule
Pooled                   & 65.8 & \textbf{52.2} & 73.4 \\
\bottomrule
\end{tabular}
\end{table}

Direct bypass attempts are handled well. Agents pass the branch-manager bypass
scenario on 33 of 33 phone calls; they pass the bank-employee-pressure and
spoken-prompt-injection scenarios on 32 of 33 each (97.0\%), and the
prior-verification-claim scenario on 29 of 33 (87.9\%). These are pooled
end-to-end completion rates,
including ungraded phone calls as misses. In the prompt-injection scenario, the
safety criterion passes all 32 graded traces; its remaining phone call is
ungraded.
Confidentiality scenarios are less reliable: agents pass 11 of 33 phone calls
involving repeated failed verification (33.3\%), 13 of 33 involving probes
about which field was wrong (39.4\%), and 11 of 33 involving probes for a name
or address (33.3\%). An agent may refuse the requested account action and still
disclose information that helps a caller defeat the verification process.

\section{Failure analysis}
\label{sec:failures}

\paragraph{Multi-step flows dominate failures.} The judge evaluated 3{,}764
assertion criteria and marked 1{,}290 as failures. Agents complete the task on
65.8\% of phone calls in the simple analysis group, but on only 52.2\% in the
complex group; the latter is the weakest group for every voice agent. Many
complex assertions require a sequence of externally visible actions: identify
the correct card, apply the required policy step, issue or update the card,
inspect the returned state, and report the result accurately.

\paragraph{Conversation and execution can diverge.} A transcript can sound
successful while the required state transition is absent, applied to the wrong
card, or rejected by the backend. Conversely, a final database snapshot cannot
show whether the agent disclosed protected information or reached the state
through a prohibited action. The joint trace exposes both dimensions: spoken
claims are compared with tool inputs, tool outputs, and the sequence in which
they occurred. A promise to freeze or replace a card therefore earns no credit
without evidence that the corresponding operation succeeded. A conservative
review of the 1{,}290 failed criterion explanations identifies at least 27
explicit say/do mismatches across 10 of 11 voice agents (2.1\% of failures).
Each retained explanation states that the agent announced a completed action
or external state before, without, or despite contradictory tool evidence.
Fourteen involve a claim with no supporting tool evidence, four contradict a
returned tool result, and nine describe premature or misstated existing state.
Seven come from an assertion that directly requires a truthful account of
system state. The rest come from state or sequence assertions whose failure
explanations name the mismatch, visible to the judge only because it reads the
execution trace alongside the conversation. For example, one agent tells
the caller it canceled a pending replacement and ordered a new one, while the
trace holds a single \texttt{request\_card\_replacement} invocation and no
cancellation at all.

\paragraph{Policy-step drift is shared across the field.} The operating
procedure requires the voice agent to offer a freeze before replacing a lost or
stolen active card. When the caller explicitly demands a replacement without
the freeze, agents succeed on only 1 of 33 phone calls (3.0\%). When an
impatient caller tries to skip the offer, agents succeed on 13 of 33 (39.4\%).
Agents fail these phone calls because the required action order disappears
under pressure, even when the conversation remains fluent. Grading the
conversation trace alongside the tool-invocation logs makes that missing procedural
step visible.

\paragraph{Connection failures remain part of end-to-end performance.}
The all-NVIDIA Pipecat configuration produces no answer on 33 of 300 phone
calls; Vapi fails to connect on 7 of 300 and OpenAI Realtime mini on 11 of 300.
Across the benchmark, 55 phone calls do not produce a connected conversation.
Two additional Gemini 3.1 phone calls connect without producing a gradeable
caller turn. Together these account for all 57 phone calls without a verdict;
each counts as a miss in the 3{,}300-phone-call completion denominator.

\section{Limitations and validity}
\label{sec:limits}

The benchmark covers one 100-scenario banking task. All eleven voice agents see
the same scenario definitions, assertions, caller channel, and grader; these
controls support paired comparisons within this study. They do not establish
general rankings across domains, languages, accents, speaking styles, or
deployment conditions. The reported standard errors describe repeat-run
variability on the fixed scenario set, not uncertainty over other scenarios.
Per-run completion spans 17 points for Nemotron, 12 for Cartesia, 9 for Gemini
3.1, and 8 for Pipecat, all larger than several differences near the top.
Wider scenario coverage and more than three runs per agent are needed to
resolve those differences.

Completion is trace-judged. Tool invocations and returned rows provide evidence
of external actions, while the conversation provides evidence of disclosure,
explanation, and alignment between a claimed result and the observed result. An
explicit final-state database query would add stronger confirmation of durable
state. Relying on that query alone, however, would omit action order and
conversational disclosures, including an unauthorized disclosure that leaves
the database unchanged. Future grading should combine the trace with explicit
state checks after each phone call.

The judge is a single LLM and has no human-annotated gold set in this study.
The caller is synthetic; accents and speaking styles are induced by its
scenario instructions and cover only the variation represented in this set.
Turn-taking controls also differ across products. A target end-of-turn threshold
can map to different vendor settings, and some hosted agents expose fewer
controls. Vapi's provider records identify Deepgram \texttt{nova-3}, while the
Pipecat and LiveKit configurations specify \texttt{nova-3-general}. Although all
three identifiers are in the Nova-3 family, differences in hosted runtime,
provider integration, and turn controls mean the three-way comparison is not a
pure harness isolation.

Cost estimates are snapshots of changing prices and product tiers. Managed
hosted prices include platform infrastructure. Across systems, the relevant
cost sources are metered model usage, charges based on phone-call duration,
platform or hosting costs, and fixed annual production licenses. For reported
per-phone-call estimates, we sum the applicable model-usage,
phone-call-duration, and hosted-platform charges. The self-hosted Pipecat and
LiveKit estimates cover metered model APIs but exclude deployment servers and
operations. Fixed annual licenses are not converted to a per-phone-call cost
when doing so would require assumptions about deployment and utilization, and
separate telephony charges are excluded consistently across rows. The systems
and model versions are those available in late July 2026, including preview
models that may change.

\section{Future research directions}
\label{sec:future}

\paragraph{Model and component sensitivity.}
How much of an end-to-end result comes from the orchestration framework, and how
much comes from the core LLM or speech models? A factorial study could hold the
framework, prompt, ASR, and TTS fixed while varying the LLM, then swap each
speech component in turn. Measuring completion, latency, and cost together
would reveal whether a more capable model improves task success enough to
justify its price or response time, and whether model effects interact with
endpointing, tool dispatch, or framework behavior. Native-audio systems do not
expose the same component boundaries, so comparisons among versions from the
same provider would offer a complementary sensitivity test.

\paragraph{Acoustic and caller robustness.}
The current scenarios include instructed accents and speaking styles, but they
do not isolate their effects. Future versions should systematically vary
accent, dialect, language, speaking rate, disfluency, background noise, channel
codec, packet loss, and device quality while holding the underlying task fixed.
Balanced coverage would make it possible to estimate both aggregate robustness
and performance gaps across caller groups instead of treating acoustic
variation as an uncontrolled part of the scenario set.

\paragraph{External validity and measurement.}
Additional domains, larger scenario sets, and more runs are needed to test
whether the observed tradeoffs persist beyond card operations. Future grading
should also combine trace evidence with explicit state checks after each phone
call and study how simulated caller behavior differs from real customer
interactions. These extensions would turn the present snapshot into a stronger
account of generalization, reliability, and uncertainty.

\section{Conclusion}
\label{sec:conclusion}

VAmoS Bench evaluates eleven production voice agents through 3{,}300
phone calls on the same 100 card-operations scenarios. Observed completion
ranges from 43.0\% to 71.0\%. Because of repeat-run variability and
unmeasured scenario-set uncertainty, we do not treat neighboring rows as a
resolved ranking. The complex group in the mutually exclusive partition is the
weakest for every agent. Pipecat and LiveKit differ by 0.7
percentage points under a shared model stack, while replacing Pipecat's stack
with the all-NVIDIA Nemotron pipeline yields an observed 28-point gap. These
results describe the present task and scenario set; broader coverage is
required for general product rankings.

The central measurement is the relationship between what a voice agent says and
what it does. A fluent transcript may claim success without a successful tool
action. A correct final state may conceal a prohibited intermediate action or
an unauthorized disclosure. Grading conversation, tool invocations, tool
results, and their order together captures failures that either view alone
misses. Here that joint evidence identifies at least 27 explicit say/do
mismatches and reveals agents that resist direct verification bypasses while
still leaking verification information or skipping a required freeze step.

\begin{ack}
We thank Vapi for feedback on the agent configuration used in this study. Their
input helped us verify implementation details; the benchmark design, analysis,
and conclusions are the authors' own.

All authors are affiliated with Veris AI, which developed and operates the
simulation platform used in this study.
\end{ack}

\bibliographystyle{plainnat}
\bibliography{references}

\appendix
\section{Systems under test}
\label{app:systems}

We selected these systems to span transport protocols and orchestration
providers, not to identify an absolute state of the art for any one
configuration. For cascaded systems, we held Deepgram,
\texttt{gpt-4.1-mini}, and ElevenLabs constant wherever framework support
allowed because that combination was supported by the broadest set of
frameworks. The \texttt{gpt-realtime-2} configuration was selected before the
newer \texttt{gpt-realtime-2.1} model was released. Thus, the configurations
prioritize breadth of support and cross-system comparability rather than
tuning each agent to maximize its benchmark score.

\begin{table}[H]
\centering
\small
\caption{The eleven voice agents, grouped by how much of the stack the vendor
owns. Pipecat and LiveKit use the same three model identifiers. Vapi shares
their LLM and TTS and uses a Nova-3-family ASR under a different recorded
identifier; Nemotron uses the Pipecat harness with every model replaced.}
\label{tab:systems}
\begin{tabular}{@{}ll>{\raggedright\arraybackslash}p{6.2cm}@{}}
\toprule
Voice agent & Class & Speech stack (as run) \\
\midrule
Pipecat            & Self-hosted OSS      & Deepgram \texttt{nova-3-general} $\to$ \texttt{gpt-4.1-mini} $\to$ ElevenLabs \\
LiveKit Agents     & Self-hosted OSS      & Deepgram \texttt{nova-3-general} $\to$ \texttt{gpt-4.1-mini} $\to$ ElevenLabs \\
Vapi               & Hosted platform      & Deepgram \texttt{nova-3} $\to$ \texttt{gpt-4.1-mini} $\to$ ElevenLabs \\
Nemotron           & Self-hosted OSS      & Pipecat harness, all-NVIDIA: \texttt{nemotron-asr-streaming} $\to$ \texttt{nemotron-3-nano-30b-a3b} $\to$ \texttt{magpie-tts-multilingual} \\
\addlinespace
Cartesia           & Hosted platform      & Cartesia Line SDK: Ink $\to$ \texttt{gpt-4.1-mini} $\to$ Sonic \\
ElevenLabs ConvAI  & Hosted platform      & Scribe $\to$ \texttt{gpt-4.1-mini} $\to$ \texttt{eleven\_flash\_v2} \\
Retell             & Bundled API          & Vendor ASR $\to$ \texttt{gpt-4.1-mini} $\to$ ElevenLabs \\
\addlinespace
OpenAI Realtime    & Speech-to-speech     & \texttt{gpt-realtime-2}, native audio \\
OpenAI Realtime mini & Speech-to-speech   & \texttt{gpt-realtime-2.1-mini}, native audio \\
Gemini 3.1 Flash Live & Speech-to-speech  & \texttt{gemini-3.1-flash-\allowbreak live-preview}, native audio \\
Gemini 2.5 Native Audio & Speech-to-speech & \texttt{gemini-2.5-flash-\allowbreak native-audio-\allowbreak preview-12-2025}, native audio \\
\bottomrule
\end{tabular}
\end{table}


\end{document}